\documentclass[iicol,pdflatex,sn-mathphys-num]{sn-jnl}

\usepackage[english]{babel}
\usepackage{graphicx}%
\usepackage{multirow}%
\usepackage{amsmath,amssymb,amsfonts}%
\usepackage{amsthm}%
\usepackage{mathrsfs}%
\usepackage[title]{appendix}%
\usepackage{xcolor}%
\usepackage{textcomp}%
\usepackage{manyfoot}%
\usepackage{booktabs}%
\usepackage{algorithm}%
\usepackage{algorithmicx}%
\usepackage{algpseudocode}%
\usepackage{listings}%
\usepackage{dsfont}
\RequirePackage[separate-uncertainty=true,list-final-separator={, and }]{siunitx}
\usepackage[version=4]{mhchem}
\usepackage{hyperref}
\usepackage[nameinlink, capitalize]{cleveref}
\usepackage{cuted} 
\usepackage{comment}
\usepackage{subfigure}
\usepackage{xspace} 
\let\orcidlogo\relax 
\usepackage{orcidlink} 

\crefname{figure}{Figure}{Figures}
\Crefname{figure}{Figure}{Figures}
\crefname{equation}{Equation}{Equations}
\Crefname{equation}{Equation}{Equations}
\crefname{table}{Table}{Tables}
\Crefname{table}{Table}{Tables}
\crefname{section}{Section}{Sections}
\Crefname{section}{Section}{Sections}
\newcommand{\minifool}{\texttt{MiniFool}\xspace}

\begin{document}

\title[\minifool]{\minifool: Physics-Constraint-Aware Minimizer-Based Adversarial Attacks in Deep Neural Networks}

\author[1,2]{\fnm{Lucie} \sur{Flek}\orcidlink{0000-0002-5995-8454}}

\author[3,6]{\fnm{Oliver} \sur{Janik}\orcidlink{0009-0007-3121-2486}}

\author[4]{\fnm{Philipp Alexander} \sur{Jung}\orcidlink{0000-0002-2511-1490}}

\author[1,2]{\fnm{Akbar} \sur{Karimi}\orcidlink{0000-0002-5132-2435}}

\author[5]{\fnm{Timo} \sur{Saala}\orcidlink{0009-0003-4401-3970}}

\author[4]{\fnm{Alexander} \sur{Schmidt}\orcidlink{0000-0003-2711-8984}}

\author[5]{\fnm{Matthias} \sur{Schott}\orcidlink{0000-0002-4235-7265}}

\author[3]{\fnm{Philipp} \sur{Soldin}\orcidlink{0000-0003-1761-2495}}

\author[3,7]{\fnm{Matthias} \sur{Thiesmeyer}\orcidlink{0009-0003-0005-4762}}

\author[3]{\fnm{Christopher} \sur{Wiebusch}\orcidlink{0000-0002-6418-3008}}

\author[4]{\fnm{Ulrich} \sur{Willemsen}\orcidlink{0009-0006-5504-3042}}

\affil[1]{\orgdiv{\small Bonn-Aachen International Center for Information Technology}, 
\orgname{University of Bonn},
\orgaddress{\postcode{53115} \city{Bonn}, 
\country{Germany}}}

\affil[2]{\orgdiv{\small Lamarr Institute for Machine Learning and Artificial Intelligence}, 
\orgname{North Rhine-Westphalia},
\country{Germany}}

\affil[3]{\orgdiv{\small III. Physikalisches Institut B}, \orgname{RWTH Aachen University}, \orgaddress{\postcode{52056} \city{Aachen}, \country{Germany}}}

\affil[4]{\orgdiv{\small III. Physikalisches Institut A}, \orgname{RWTH Aachen University}, \orgaddress{\postcode{52056} \city{Aachen}, \country{Germany}}}

\affil[5]{\orgdiv{\small Physikalisches Institut}, 
\orgname{University of Bonn},
\orgaddress{\postcode{53115} \city{Bonn}, 
\country{Germany}}}

\affil[6]{\small \orgdiv{now at} \orgname{Erlangen Centre for Astroparticle Physics, Friedrich-Alexander-Universität Erlangen-Nürnberg, 91058 Erlangen}, \country{Germany}}

\affil[7]{\small \orgdiv{now at} 
\orgname{Dept. of Physics and Wisconsin IceCube Particle Astrophysics Center, University of Wisconsin—Madison}, 
\country{USA}}

\abstract{In this paper, we present a new algorithm, \minifool, that implements physics-inspired adversarial attacks for testing neural network-based classification tasks in particle and astroparticle physics. While we initially developed the algorithm for the search for astrophysical tau neutrinos with the IceCube Neutrino Observatory, we apply it to further data from other science domains, thus demonstrating its general applicability. Here, we apply the algorithm to the well-known MNIST data set and furthermore, to Open Data data from the CMS experiment at the Large Hadron Collider. The algorithm is based on minimizing a cost function that combines a $\chi^2$ based test-statistic with the deviation from the desired target score. The test statistic quantifies the probability of the perturbations applied to the data based on the experimental uncertainties. For our studied use cases, we find that the likelihood of a flipped classification differs for both the initially correctly and incorrectly classified events. When testing changes of the classifications as a function of an attack parameter that scales the experimental uncertainties, the robustness of the network decision can be assessed. Furthermore, this allows testing the robustness of the classification of unlabeled experimental data.}




\maketitle



\section{Introduction}

In recent decades, machine learning-based algorithms have become a central cornerstone of data analysis in particle and astroparticle physics~\cite{Erdmann:2021jbm}. Many current results have become possible because of the use of deep neural networks in classification and regression tasks. Prime examples are several recent discoveries from the IceCube Neutrino observatory~\cite{IceCube:2024nhk,IceCube:2023ame,IceCube:2022der} as well as unprecedented composition measurements from the Pierre Auger Observatory~\cite{PhysRevLett.134.021001}, and enhanced detector simulations of complex detectors at the Large Hadron Collider (LHC)~\cite{hashemi2024ultra,lai2022explainable,lagan}.

Given the importance of enhanced precision in data analysis, understanding network decisions and their robustness with respect to subtle changes of the input becomes highly important. It is well-known in machine learning that adversarial attacks that modify the input with imperceptible perturbations can drastically change the output of networks~\cite{xu2020adversarial}. This is highly relevant for the safety of networks in commercial and financial applications. However, for particle physics applications, adversarial attacks turn out to be valuable tools as well. Here, one often considers classification tasks, where neural networks split a data set of discretely recorded events into signal and background.
The training is usually based on labeled Monte Carlo simulations
of such events~\cite{deFavereau:2013fsa}. Subtle mismodeling of the experimental instruments can cause the networks to learn features that might not be real, therefore degrading the classification performance, once applied to the experimental data.
This degradation is difficult to quantify or might not even be recognized. In view of these challenges, the application of adversarial attacks as a tool allows not only for testing the robustness of network decisions~\cite{carlini2017} but also enables the improvement of the robustness itself by specifically training the networks with adversarially attacked data~\cite{Goodfellow:2014rpb}.

In this paper, we present a new algorithm, \minifool, 
that has been developed in the context
of the search for astrophysical tau neutrinos~\cite{IceCube:2024nhk} in the IceCube Neutrino Observatory~\cite{IceCube:2016zyt}. Most well-known attack algorithms in the literature focus on perturbing the input data in a minimal way, but they often do not obey experimental uncertainties or boundary conditions, such as conservation laws of physics
or non-physical data values, such as negative signals.
Even when supplementing the 
attack algorithm with additional boundary constraints, the 
minimal change in data that leads to a change in classification 
might not represent a data event within the physically allowed phase space. Consequently, the statistical and physical interpretation of the success rate of attacks is difficult.

On the contrary, the approach of \minifool is to optimize the attack also with respect to physically reasonable changes. 
This is based on the minimization of a cost function that balances the required change of the target score with the required change of the input. This change is quantified by the mean square difference relative to the physical uncertainty of the data.
As a result, we can use the success rate of the attack to assess the robustness of the network. 
If the perturbation of an attack that is required to reach a changed target classification becomes too costly with respect to the data uncertainties, \minifool will not change the classification.

In this paper, we start by briefly reviewing existing attack algorithms and their respective difficulty when applying them to our use cases. The underlying test statistic of \minifool is motivated next.
We then present the results for an initial example case, where we apply the algorithm to the MNIST data set~\cite{lecun2010mnist,deng2012mnist} --- a well-known classification challenge in the literature.
We then turn to the identification of tau neutrinos in IceCube~\cite{IceCube:2024nhk}, the context within which this algorithm was developed.
Finally, we study the event classification of b-quark-jets using CMS open data from the Large Hadron Collider \cite{bib.open_data}. Here we see that the algorithm is generalizable to other particle physics examples.
The paper ends with a discussion of the results.


\section{Related Work}


Adversarial attacks have been studied for more than a decade and are evolving as a valuable tool in astroparticle and elementary particle physics (See e.g.~\cite{Flek:2025ecg,Stein:991721,Stein:2022nvf}). 
Several established adversarial attacks have been tested in those fields.
A very well-known method is the \emph{Fast Gradient Sign Method (FGSM)}~\cite{Goodfellow:2014rpb}.
The algorithm evaluates the signed gradient of the cost function with respect to the input data for constructing a minimally perturbed input.
Smaller perturbations can be achieved with the \emph{Projected Gradient
Descent (PGD)} ~\cite{Madry:2017tvh}
or the \emph{DeepFool} algorithm~\cite{Deepfoolpaper}. 
These can be understood as iterative extensions of the FGSM algorithm that more efficiently find a minimally perturbed solution to a changed classification.
Used norms for the cost function
vary between $L_\infty$, $L_2$, and $L_1$.

This approach makes these algorithms particularly well-suited for generating adversarial examples with changes that are imperceptible to human perception.
However, what is common in the aforementioned algorithms is that these minimal perturbations of the input data are not constrained, and non-physical or negative pixel values are possible,
if this improves the solution.
Unless the algorithms are supplemented with counter-measures by additional constraints on the ranges of possible
changes of the input data (See e.g.~\cite{masterjanik}), the applicability to classification tasks in particle physics analyses is limited.

The approach of \minifool is different. Instead of minimizing the distance metric to the decision boundary, the cost function of perturbations includes a distance metric that quantifies the credibility of perturbations given by the uncertainties of the input data. By this, we minimize the perturbation with respect to maximum plausibility. A very similar approach has been found in the L-BFGS algorithm by Szegedy et al.~\cite{szegedy2013intriguing}
where they minimize a scaled distance-metric $ ||r||_2 $ of the unperturbed $\vec{x}$ and perturbed $\vec{x}+\vec{r}$  input plus a loss term $f $
that depends on the perturbed data and the cross-entropy loss of the targeted classification label $l$
\begin{equation}
  \min_r \ [ s\cdot ||\vec{r}||_2 + f(\vec{x}+\vec{r} ,l) ] ~.
\end{equation}
This minimization is performed iteratively with different parameters $s>0$ to find a global optimum of minimum distance. This concept is adopted by Carlini et al.~\cite{carlini2017}, where different loss functions $f$ are used for evaluating the robustness of network classifications.
In our approach, we deviate from a simple scale parameter $s$ but add physical uncertainties to the distance metrics. The scale of the attack $s$ is applied to these uncertainties. 
Furthermore, instead of evaluating the loss function of the network,
we replace it with a statistically motivated $L_2$ distance metric of the perturbed output and a targeted label score.



\section{Minimizer-Based Network Perturbations}




In this work, we restrict ourselves to a classification task
with $m$ classes.
This means that the network output $\vec{f}(\vec{x};\vec{\theta}) $ is a vector of dimension $m$ which is a non-linear function of the input data $\vec{x} $.
Here $\vec{\theta} $ are the hidden parameters of the network. It is
very common to use a softmax activation function in the final output layer, which results in output data that can be interpreted such that the value $f_i \in \vec{f}$ is the probability of the input belonging to class $i\in m$.
The loss function during training is usually the cross-entropy
and the maximum 
\begin{equation}
 i^* = \underset{i}{\arg\max} f_i (\vec{x};\vec{\theta}) 
\end{equation}
 becomes the assigned class.
The goal of adversarial perturbations is to change the original input $\vec{x}^0$ to $  \vec{x}^a$ such that the output becomes
\begin{equation}
  \underset{i}{\arg\max} f_i (\vec{x}^a;\vec{\theta}) \ne i^* ~.
\end{equation}

As a distance metric for quantifying the perturbations of the input data 
$x_i^0 \in \vec{x}^0 $, we use a squared  $L_2$ norm of  mean-squared deviations of perturbed feature values $\vec{x}^{\mathrm{a}} $ with respect to the original values $\vec{x}^{\mathrm{0}}$, 
\begin{equation}
  \eta = \frac{1}{N}\sum_{i=1}^{N}\left(\frac{x_i^0 - x_i^{\mathrm{a}}}{\sigma_i}\right)^2 ~.
  \label{equation.eta}
\end{equation}
The deviation of each feature is individually normalized to its uncertainty $\sigma_i$, similar to a $\chi^2 $ value in statistics, and the function is overall normalized to the dimension $N$ of the input data.
Note that this metric can be easily extended to correlated input features 
by using $ (\vec{x}^0 - \vec{x}^{a})^T \mathbf{\Sigma}^{-1} (\vec{x}^0- \vec{x}^{a} ) $ where $\mathbf{\Sigma } $ is the covariance matrix.

Additionally, for our classification attack, we have to define a target score $g$ for the attack. Here, we have two choices. Without loss of generality, assuming the target output of the class $f_{i^*} $ approaches $1$ for a very confident classification, a reverted classification corresponds to $f_{i^*} (\vec{x}^a;\vec{\theta}) {=} g$ with $g = 0$, or another desired score $g \ne f_{i^*}(\vec{x}^0;\vec{\theta})$. Note that this requirement does not restrict the possible values of the other $m-1$ classes $f_{i\ne i^*} $. Alternatively, one can explicitly set target scores $ g_i \in\vec{g} $ for a subset or all output classes.

In the second case, the distance of a network output $\vec{f} $ to the target score can then be defined by the $L_2 $ metric similarly to the input metric $\Delta^2$ with $\Delta = \vec{f}(\vec{x}^a;\vec{\theta}) -\vec{g} $.
In our tests, we focus on the first variant of a changed classification without requiring a specific new label $i$ or a specific output score for other labels $i \ne i^*$.
For this case, the distance for the target score simplifies to $(f_{i^*} (\vec{x}^a;\vec{\theta}) - g)^2$, where typically $g = 0$. This means we are targeting a maximal change of the initial classification without constraining the outputs for other classes.
Obviously, also a mixed metric of the two alternatives, e.g., setting multiple output classes to zero, can be implemented, if conceptually required.
For \minifool, we now construct the total metric as 
\begin{equation}
  \lambda (\vec{x}^0; \vec{x}^{a}; \vec{\theta} ) =  \eta(\vec{x}^0; \vec{x}^{a} ) + \beta \cdot (f_{i^*}(\vec{x}^{a}; \vec{\theta}) - g)^2
  \label{equation.cost_function}
\end{equation}
where $f_{i^*} $ depends on the perturbed input $\vec{x}^{a} $ and the parameters $\vec{\theta}$ of the tested network. The tunable meta-parameter
$\beta $ has the default value $1$. It allows re-weighting the relative importance of the two distance terms. A value $\beta \ne 1$ can e.g.\ be indicated if the classification  $f$ is not normalized as probability between \numrange{0}{1}.
For each individual data event, this metric is minimized with respect to the adversarial perturbations to obtain our adversarial example $\vec{x}_{min}^a$
\begin{equation}
\vec{x}_{min}^a = \vec{x}^a \quad \mbox{for} \quad 
  \underset{\vec{x}^a}{\arg \min} \ \lambda (\vec{x}^0; \vec{x}^{a}; \vec{\theta}) ~.
\end{equation}
Depending on the values of the data uncertainties $\sigma_i$
as well as the 
target score, an optimum solution is found for each individual event. If small perturbations can easily achieve the desired target score, the classification will be changed with minimal changes to the input, while if small perturbations cannot easily achieve the target, the optimum solution would favor not changing the initial classification.

The concept of a changed classification depending on the assumed uncertainty scale can be extended into a test that allows assessing the robustness of the classification of the network for each individual event. In the simplest version, we scale the nominal uncertainties $\vec{\sigma}^0 $ with a scalar parameter $s$ which we call \emph{attack parameter}
\begin{equation}
  \vec{\sigma} = s \cdot \vec{\sigma}^0 
\end{equation}
Here, $s=1$ corresponds to the nominal uncertainties.
Each individual data event can be scanned by evaluating the optimum classification score as a function of this attack parameter. A flipped classification only for $s \gg 1 $ would indicate a robust classification, while a flipped classification for $s \le 1 $ signals a 
less robust classification.
Note, that for cases where the uncertainties 
 $\vec{\sigma}$ and their correlations are not  known or imprecise, the attack parameter $s$  still provides a useful algorithmic knob for establishing a relative robustness score
 that, however, cannot be interpreted statistically.

\minifool is available as both a TensorFlow~\cite{abadi2016tensorflow,tensorflow2015-whitepaper} and a PyTorch~\cite{paszke2019pytorch,paszke2017automatic} implementation. 
The minimization is computationally more demanding than many other adversarial algorithms.
The execution time to create one adversarial example depends on the size of the image and network, as well as on the choice of the hyperparameters of the minimizer. We employ the Adam optimizer~\cite{kingma2017adammethodstochasticoptimization} with a default learning rate of $1\cdot 10^{-5}$. For parameter scans over the attack parameter $s$, we scale the learning rate to $0.1\cdot s$ to accelerate convergence. Validation tests confirmed that smaller learning rates produce identical results, though these hyperparameters should be tuned for specific applications. It can take up to a few seconds on a typical workstation and is thus slower than other adversarial attack methods. 
The code of \minifool is published as open source on GitHub~\cite{git:minifool}. A simple MNIST example is included to demonstrate its application.


\section{Warm-up: Image Recognition with the MNIST Dataset}

\begin{figure}[htbp]
  \centering
  \includegraphics[width=1.0\columnwidth]{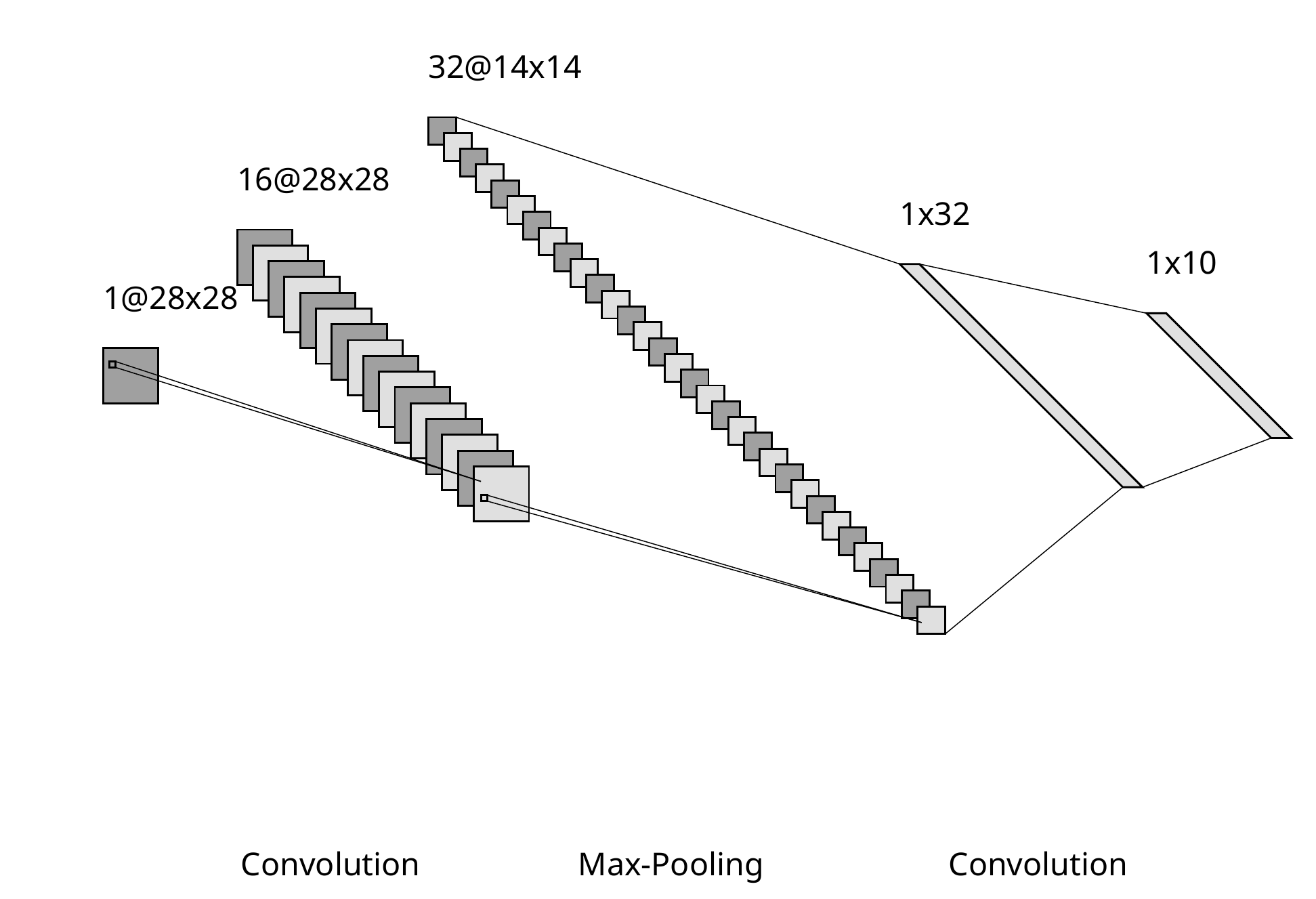}
  \caption{Simple network for the classification of MNIST images.}
  \label{fig:MNIST_Architecture}
\end{figure}

To illustrate the concept of the minimization-based adversarial attack in \minifool, we apply it to the well-known MNIST digit classification task~\cite{lecun2010mnist,deng2012mnist}. MNIST consists of grayscale images of handwritten digits from 0 to 9, each of size $ 28 \times 28$ pixels. The input dimensionality is thus $N=784$, and the number of classes is $m=10$. We train a standard feedforward neural network illustrated in~\cref{fig:MNIST_Architecture}. The model is trained using cross-entropy loss and achieves over \SI{98}{\percent} test accuracy. The trained model remains fixed throughout the adversarial attack process.

\begin{figure}[htbp]
\centering

\includegraphics[width=.99\columnwidth,trim={80 0 80 0},clip=true]{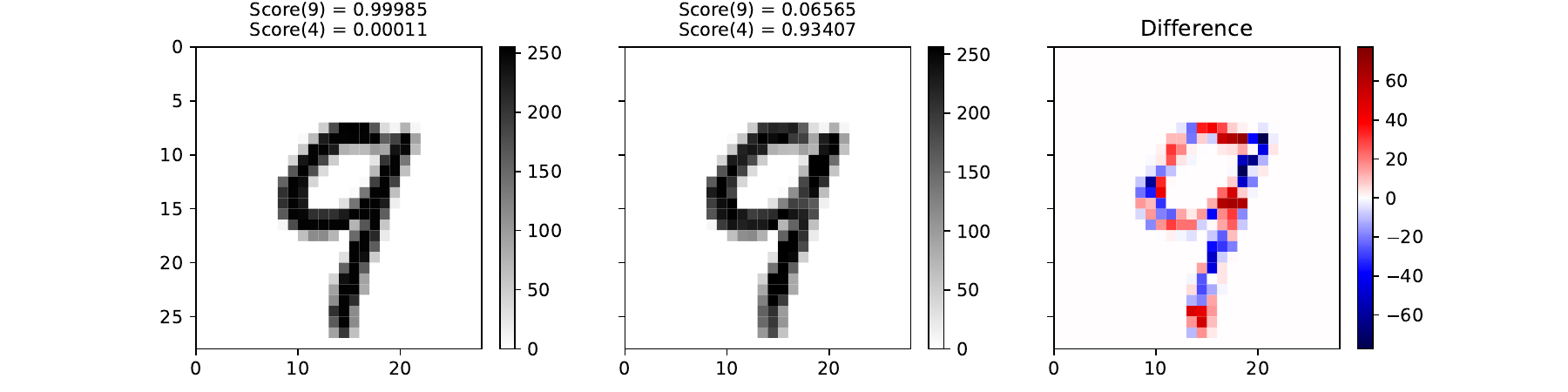}\\
\includegraphics[width=.99\columnwidth,trim={80 0 80 0},clip=true]{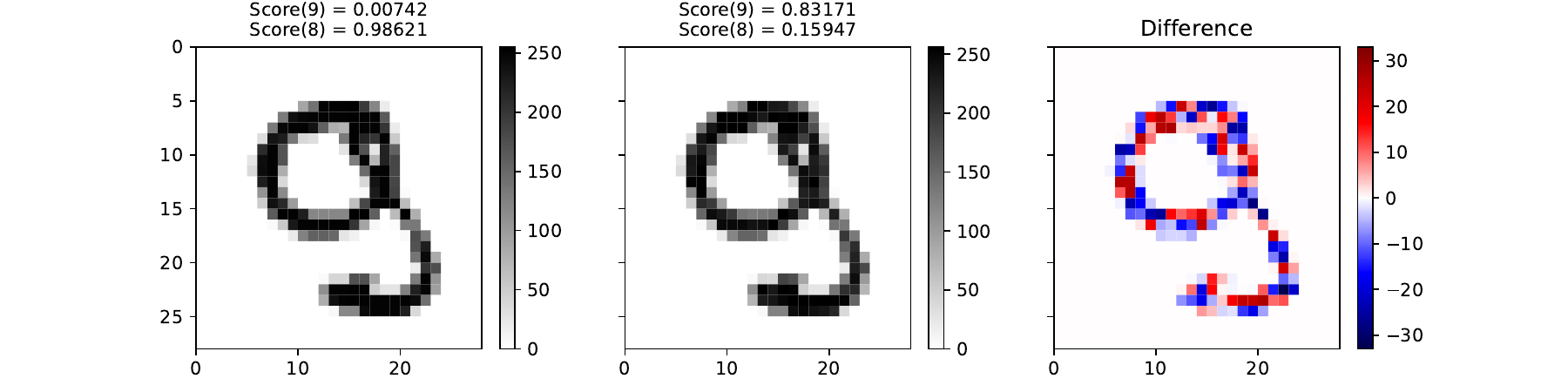}
\caption{Example of \minifool Attacks on MNIST images. 
The upper row shows a correctly and the bottom row a wrongly classified "9". The bottom row is wrongly classified as an "8".
Left are the original images, the middle shows the images perturbed by \minifool with an attack parameter $s=0.2 $. The right figures show the applied optimal perturbations. The respective classification scores can be seen above the images. 
}
\label{fig:MNIST_Classification}
\end{figure}

We apply the \minifool attack to two images illustrated in~\cref{fig:MNIST_Classification}. The first row shows an image of a "9" that is correctly classified as "9". After the \minifool attack, it is classified as "4" with a softmax score of \SI{93}{\percent} confidence, while the score for "9" is reduced to 
 \SI{6}{\percent}.
The second row shows a true "9" that is wrongly classified as "8" with a score of almost \SI{99}{\percent} 
while the score of the true label "9" is less than \SI{1}{\percent}. 
After the attack, which only targets changing the "8" into something else, it is correctly classified as "9" with a score of \SI{83}{\percent}, and the score for "8" is reduced to \SI{16}{\percent}.
The respective left plot shows the original, the center the attacked, and the right the difference between the two. 
All attacks were performed with 
a simplified uncertainty model using
an attack parameter $s=0.2$ and $\sigma_i^0=x_i^0 $. During the attack, the image was not constrained to the original image's grayscale range. We observe that approximately \SI{80}{\percent} of initially incorrectly classified images receive the correct label after the attack.

\begin{figure}[htbp]
  \centering
  \includegraphics[width=.99\columnwidth]{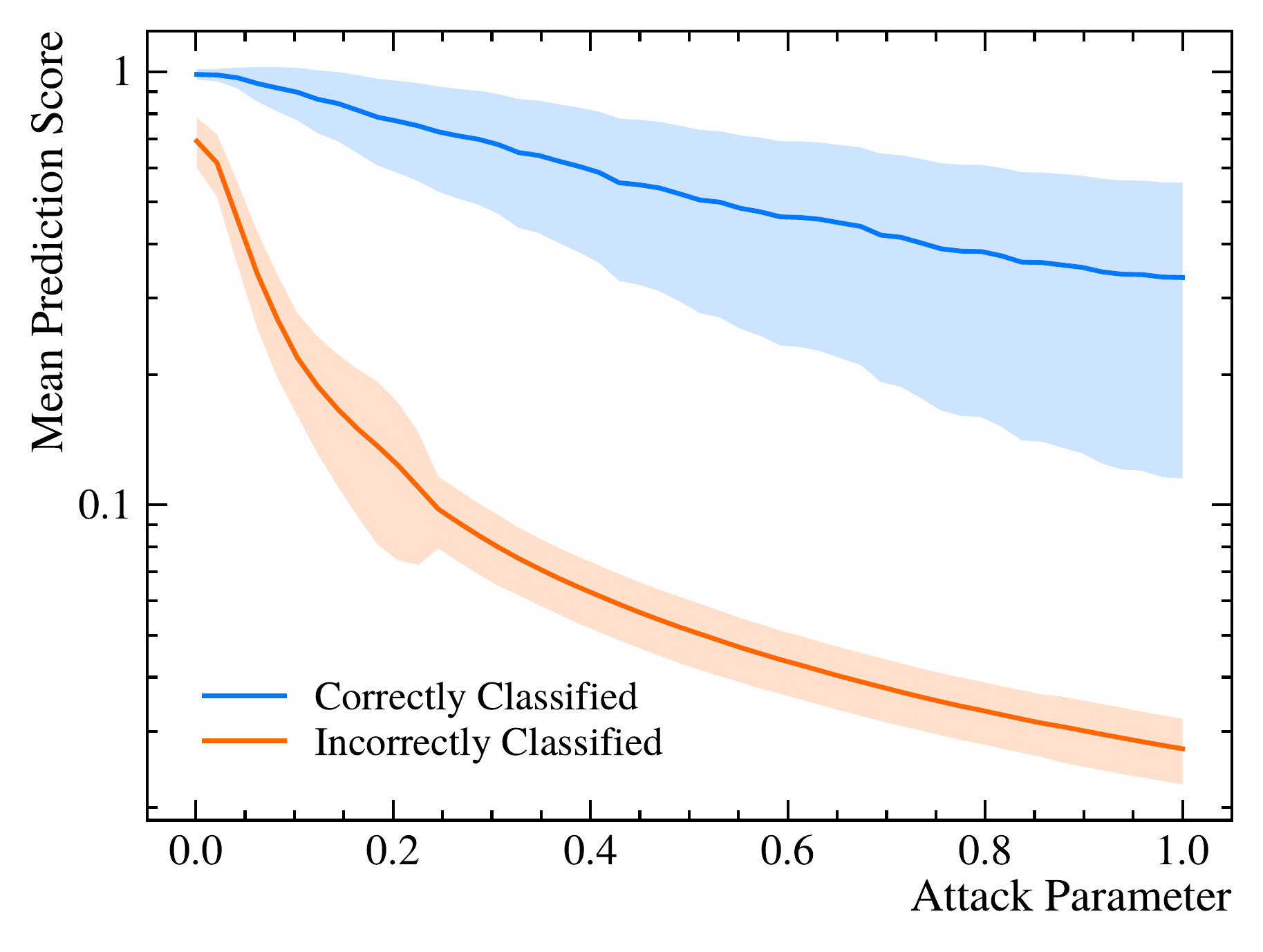}
  \caption{Average softmax score of the initially predicted class as a function of the attack parameter. The solid lines show the mean over correctly and incorrectly classified samples, while the shaded bands indicate one standard deviation. Misclassified samples exhibit a faster confidence decay under adversarial perturbation.}
  \label{fig:MNIST_attack_mean}
\end{figure}

To further analyze the behavior of the attack, we applied \minifool with varying attack parameters to two groups of test images: approximately 200 that were initially misclassified and another 200 that were correctly classified. For each group, we tracked the softmax score of the originally predicted class as a function of the attack parameter. The results, shown in~\cref{fig:MNIST_attack_mean}, display the average score of each image group with the standard deviation indicated as a shaded band. 
As expected, we observe a gradual decline in classification score with increasing attack strength.
This decline is significantly smaller for correctly classified images, while the misclassified images show a much faster drop. This indicates that incorrect predictions are typically less robust to adversarial perturbations.


\section{IceCube's Tau Neutrino Analysis}

The IceCube Neutrino Observatory~\cite{IceCube:2016zyt} is a large ice-Cherenkov detector instrumenting roughly \SI{1}{km^3} (i.e.\ \SI{1}{Gton}) natural deep ice at the geographical South Pole. It consists of 5160 optical sensors, called \emph{DOMs}, on a hexagonal arrangement of 86 vertical cables, called \emph{strings}, buried \SIrange{1450}{2450}{m} below the ice surface. The main goal of IceCube is the measurement of neutrinos from astrophysical sources. Neutrino interactions inside or close to the detector result in Cherenkov light that is detected by the DOMs. Based on the number and arrival times of detected Cherenkov photons, the direction, energy, and flavor of the neutrino can be inferred, thus establishing an astronomical telescope.

\begin{figure}[htp]
  \centering

  \includegraphics[width=0.8\linewidth]{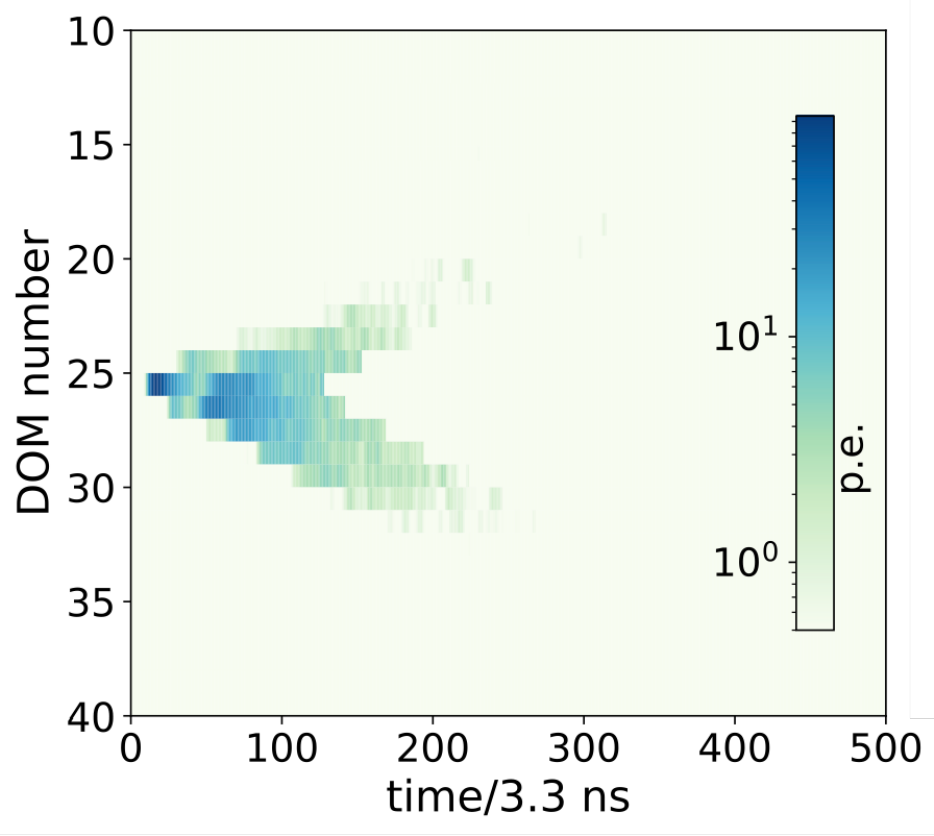}
  \caption{Example image of a recorded $\nu_\tau $ candidate event, recorded in Nov.\ 2019. The image shows the data recorded by the DOMs on the string closest to the interaction vertex (leading string). 
  The recorded DOM amplitude, normalized to photoelectrons, is represented as $60\times 500$ pixel image
  corresponding to the sensor number (or depth, respectively) along the vertical axis and the time in steps of \SI{3.3}{ns} on the horizontal axis. 
  Clearly visible is the starting point of the event, and then the distance-dependent arrival of photons at the DOMs of the string. The total number of recorded photons is \num{6000}. (modified picture taken from the supplementary material in \cite{IceCube:2024nhk})}
  \label{fig:taufrompaper}
\end{figure}

In $2013$ IceCube discovered a flux of high-energy neutrinos of astrophysical origin~\cite{IceCube:2013low}. While IceCube is sensitive to all flavors of neutrinos, $\nu_e $, $\nu_\mu$, and $\nu_\tau $, the latter tau neutrinos are particularly interesting as they are marginally produced in air showers at high energy and thus represent a unique signature of astrophysical origin~\cite{IceCube:2020fpi}. However, the clear identification of tau neutrinos is challenging for that energy range, where most astrophysical neutrinos are detected in IceCube. Tau neutrinos need to be distinguished from the other neutrino flavors $\nu_e $ and $\nu_\mu $ as well as from atmospheric muons, induced by cosmic-ray air showers, and require a challenging analysis and data selection~\cite{IceCube:2024nhk, IceCube:2020fpi, IceCube:2015vkp}.
Previous attempts to identify tau neutrinos within the recorded astrophysical neutrino events included dedicated searches for the distinct double-cascade signatures of tau neutrinos: \emph{double bang} events~\cite{Learned:1994wg}, i.e.\ two separated energy depositions, or \emph{double pulses} in raw waveforms of the DOMs and succeeded with a few candidates for $\nu_\tau$~\cite{IceCube:2020fpi, IceCube:2015vkp}. In 2024, the observatory reported the evident observation of seven events with the signature of a tau neutrino interaction in the detector, in excess of the background expectation of $0.5$ events~\cite{IceCube:2024nhk}. This analysis employed a neural network-based approach utilizing standard image recognition techniques. Here, the recorded waveforms of the DOMs, i.e.\ the recorded voltage versus time (which reflects the sensor's response to the arriving photons), are encoded as 2D images. Each image represents one of the vertical strings of 60 DOMs.
The vertical axis represents the position number of the DOM along the string, which also corresponds to the depth of the sensor. The horizontal axis is the relative time of the recorded signals in that DOM, where each pixel corresponds to a time bin of \SI{3.3}{ns}. The individual pixel values represent the measured amplitude of the waveform in units of photo-electrons. An example image is shown in \cref{fig:taufrompaper}.

A central element of the data selection of the $\nu_\tau $ search is based 
on several convolutional networks with standard VGG16 architecture~\cite{Simonyan:2014cmh} that are trained with labeled simulated data for the classification of different event types.
Inputs to these networks are each three images, corresponding to the recorded data of those three strings that are located closest to the spatial interaction vertex within the detector.
In the following, we will focus on the network that distinguishes $\nu_\tau $ induced events from $\nu_e $ events, which are the most critical background in the analysis.
The network's output score ranges between $1$ for likely $\nu_\tau $ and $0$ for likely $\nu_e $ induced events.
More details of the analysis are found in~\cite{IceCube:2024nhk}.

\begin{figure*}
\centering
 \begin{subfigure}[]
   \centering \includegraphics[width=.8\columnwidth]{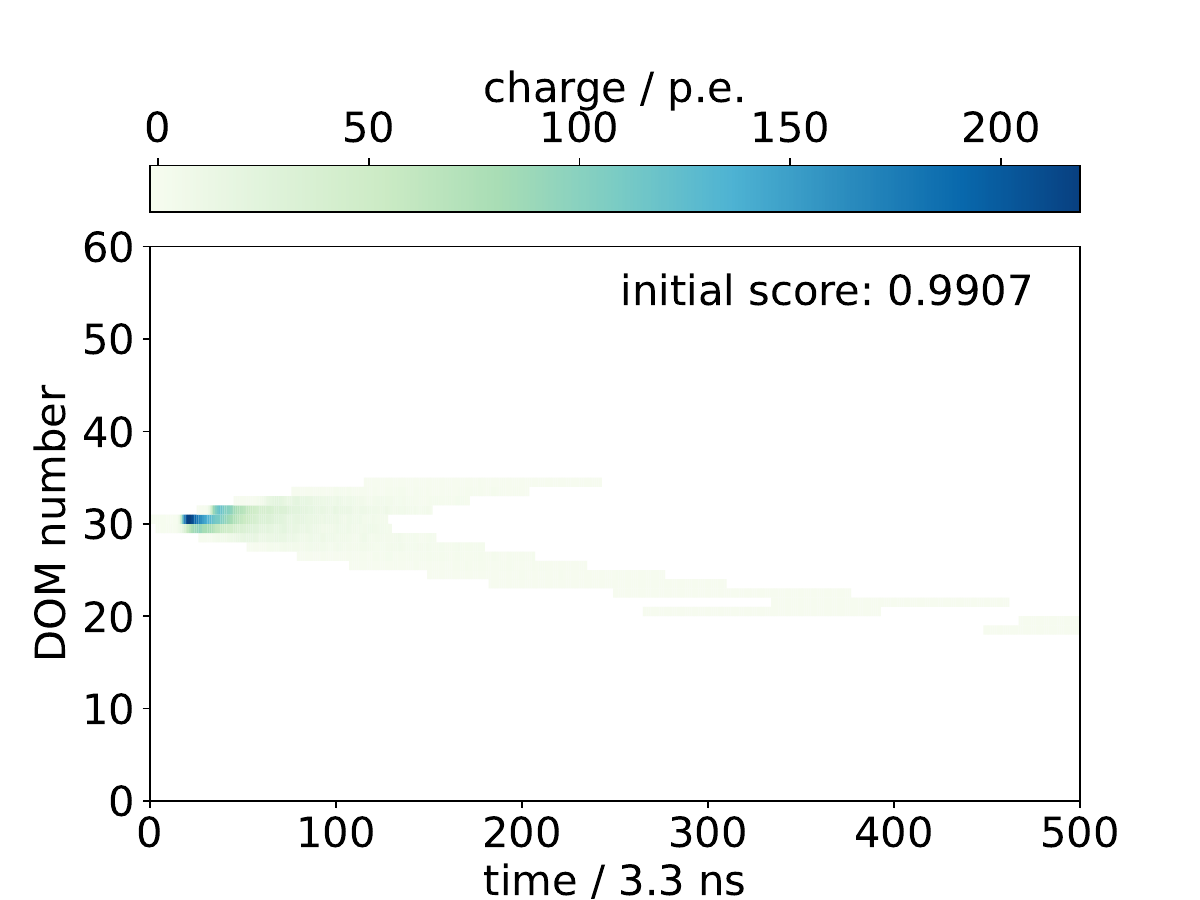}
   \label{fig:event-orig}
 \end{subfigure}
 \begin{subfigure}[]
   \centering  \includegraphics[width=.8\columnwidth]{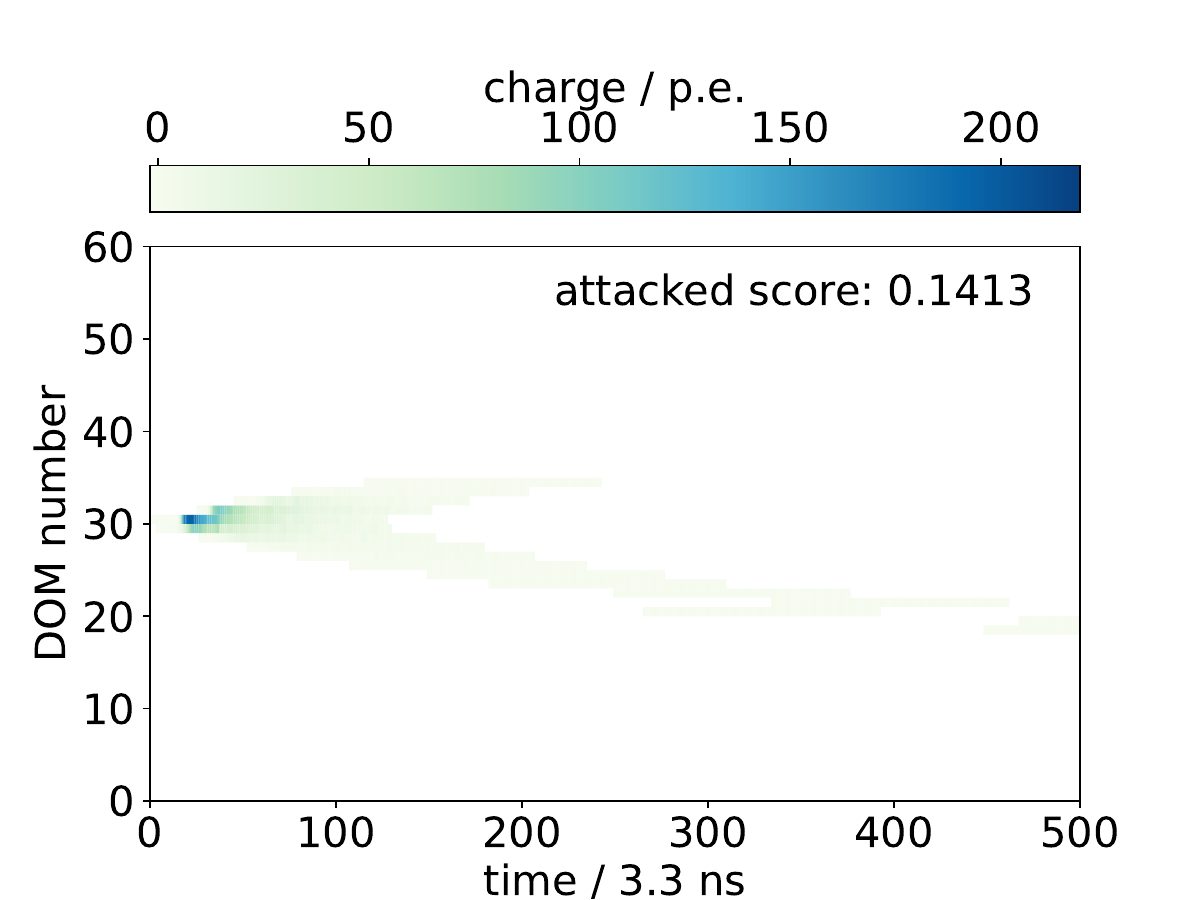}\\  
   \label{fig:event-manip}
 \end{subfigure}
 
 \begin{subfigure}[]
   \centering
   \includegraphics[width=.8\columnwidth]{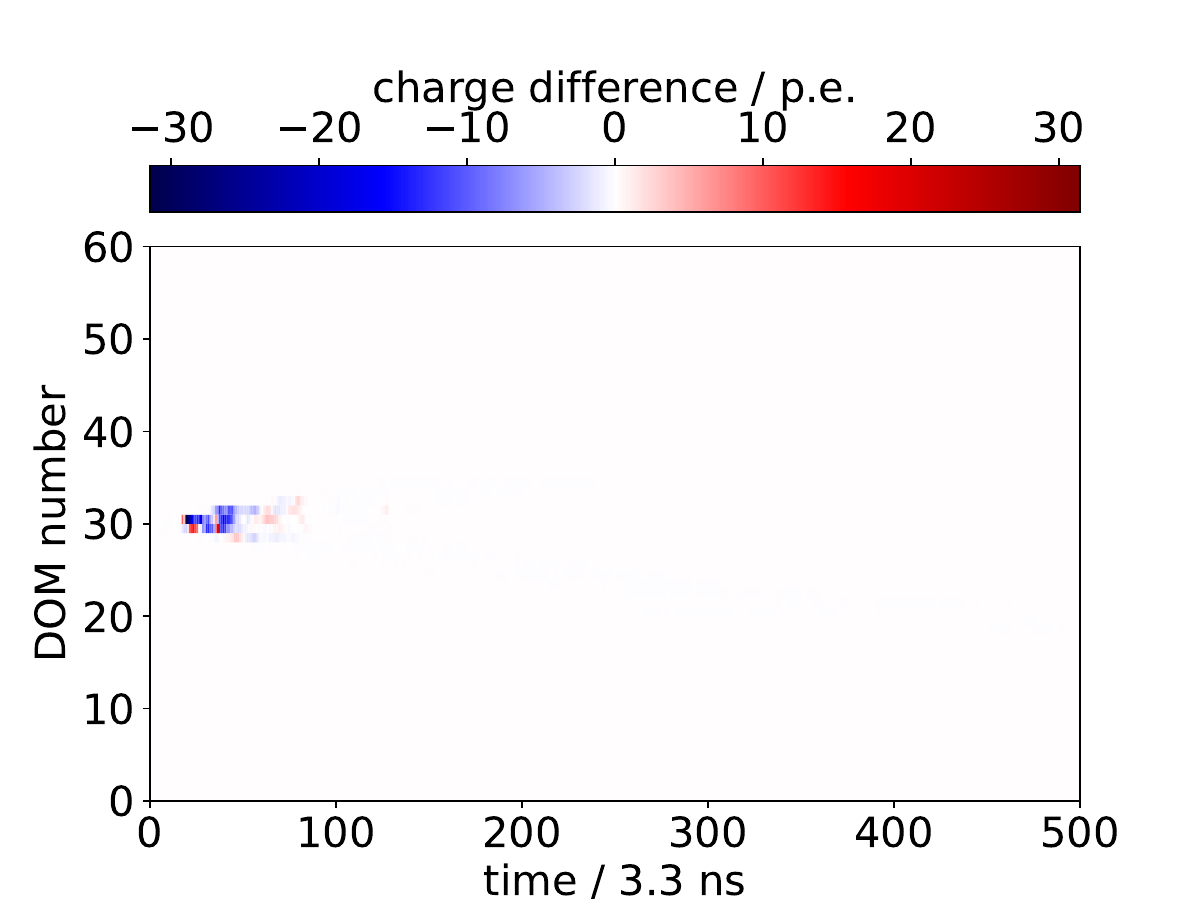}
   \label{fig:event-diff}
 \end{subfigure}
 \begin{subfigure}[]
   \centering
  \includegraphics[width=.8\columnwidth]{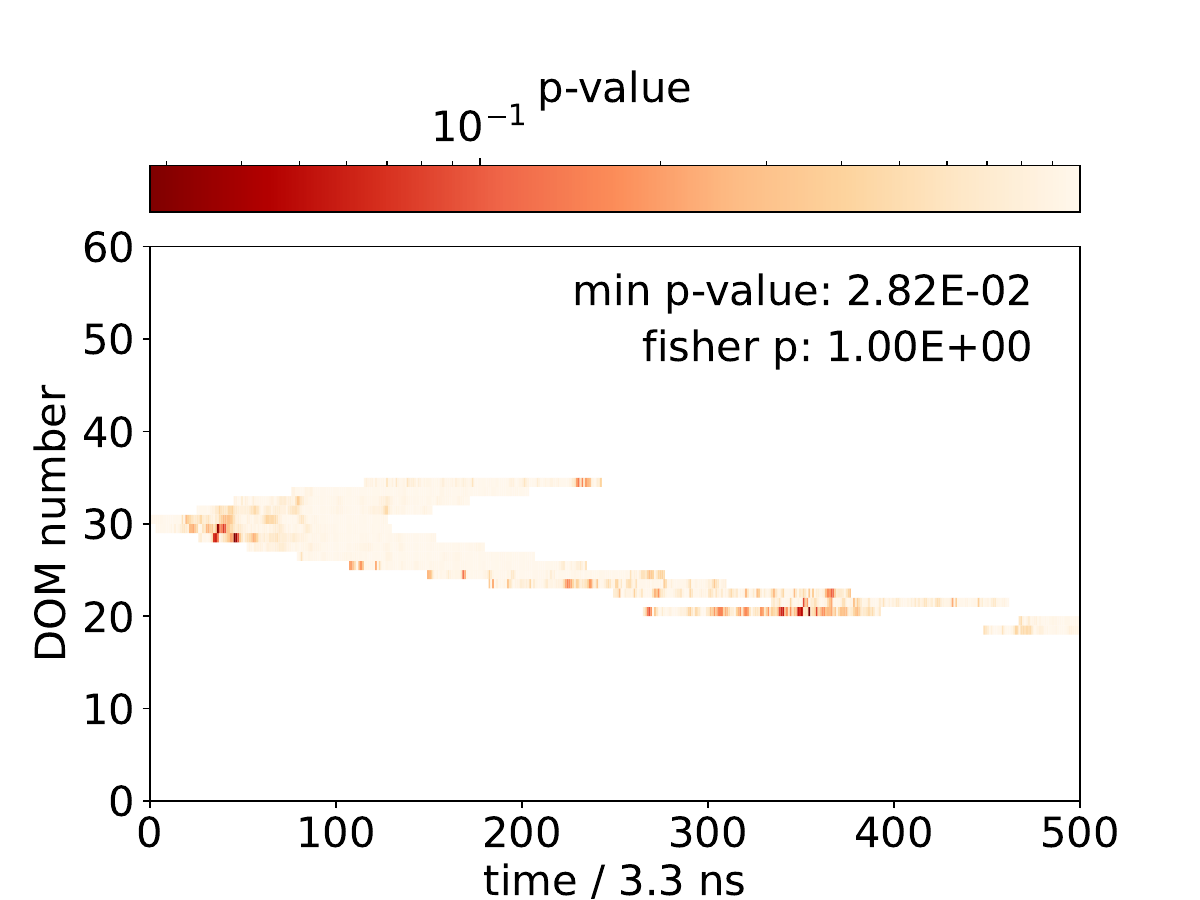}
   \label{fig:event-pull}
 \end{subfigure}

\caption{Example of an attack of a simulated IceCube event (modified images from~\cite{masterjanik}). (a) shows the original image of the measured charge versus time on the leading string. The image is attacked with \minifool and an attack parameter of $s=0.1$, resulting in (b). The difference (a)-(b) is shown in (c). Finally, (d) shows the p-value of applied changes.}\label{fig:event}
\end{figure*}

For the application of \minifool, we have chosen to attack only pixels with a non-zero recorded signal and assume the same relative uncertainty for all pixels
proportional to the recorded amplitude of that pixel.
Our default attack parameter is $ s=1 $ with 
$\sigma_i^0= 0.1 \cdot x_i^0 $.
This corresponds
to an uncertainty 
 of \SI{10}{\percent} of the respectively recorded amplitude in that pixel which roughly reflects the experimental uncertainties.
For selected $\nu_\tau $ candidates, the goal score
becomes $g=0$. More details of the implementation are given in \cite{masterjanik}.

The attack is illustrated in \cref{fig:event}, which shows a simulated $\nu_e$ event that was falsely classified as $\nu_\tau $ with a score of $f(\vec{x}^0;\vec{\theta}) = 0.9907$ (top left). After the attack (top right), the score is reduced to $f(\vec{x}^a;\vec{\theta}) = 0.1413$. The difference between the two images $\vec{x}^0- \vec{x}^{a}$ is shown on the bottom left. 
For the statistical quantification, we calculate the p-values where the applied changes are consistent with the uncertainty assumption (bottom right). This confidence calculation assumes a double-sided Gaussian distribution of width $\sigma_i^0$ for the calculation of the confidence interval of changes around the initial value $x_i^0$. The smallest p-value is \SI{2.8}{\percent}
which corresponds to 2.2 standard deviations and is thus statistically acceptable given the large number of pixels. Also, the overall p-value that we obtain by combining all p-values with Fisher's method~\cite{Mosteller01101948} is $1$.

\begin{figure*}[htbp]
\centering
\includegraphics[width=1.49\columnwidth]{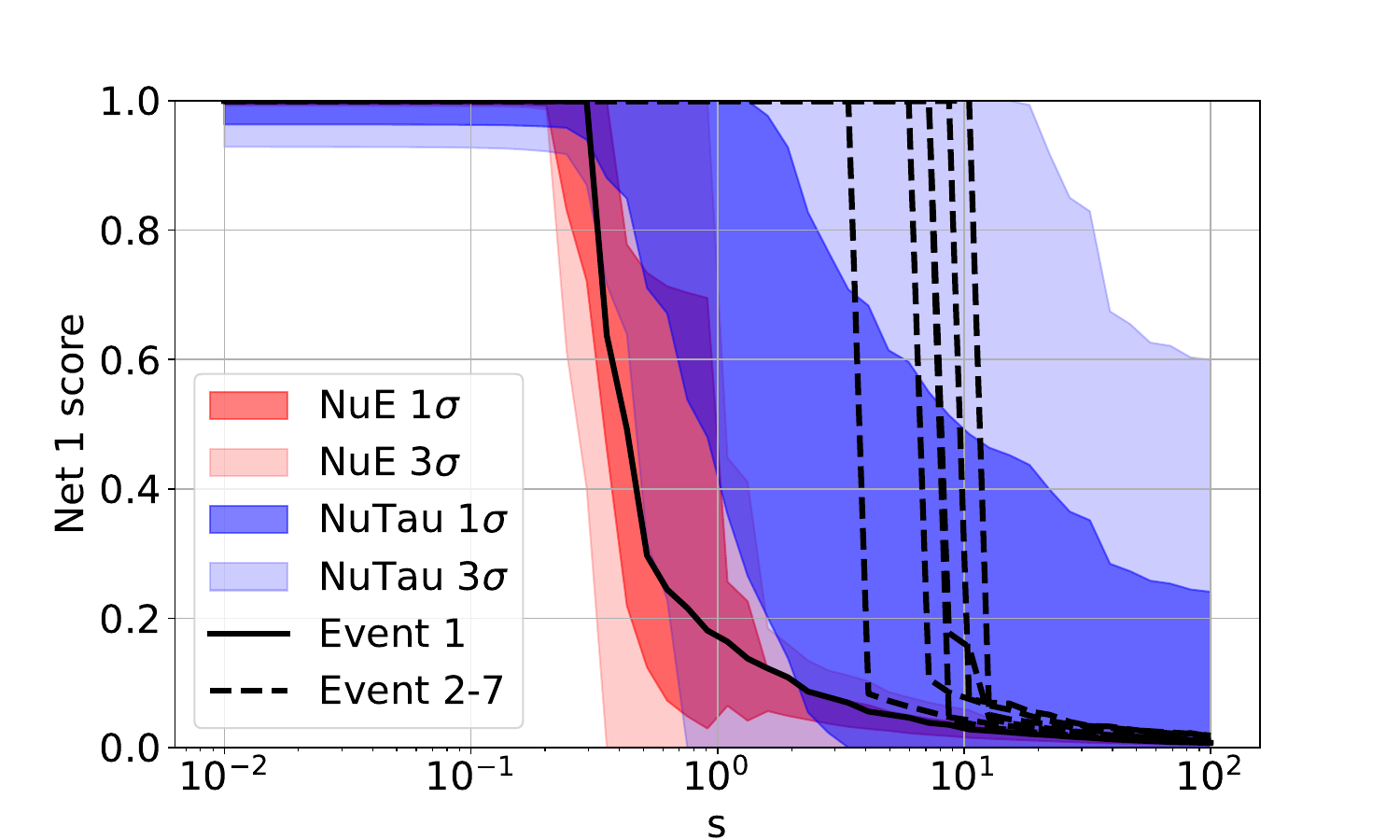}
\caption{Network score versus attack parameter\label{fig:i}. 
The picture shows the resulting network scores for \minifool attacks of the seven selected $\nu_\tau$ candidates in~\cite{IceCube:2024nhk} as black lines as a function of the assumed attack parameter $s$.
For comparison the contours show the expectation of \SI{68.3}{\percent} (dark) 
and \SI{99.7}{\percent} central quantile range for simulated signal $\nu_\tau$ events (blue) and background $\nu_e $ (red) events. 
}

\end{figure*}

For the evaluation of the seven identified $\nu_\tau $ candidate events in~\cite{IceCube:2024nhk}, we compare the application of \minifool to these events with the expectation from simulated $\nu_\tau$ and
$\nu_e$ passing the final selection.
The result of attacks is shown in \cref{fig:i} as a function of the attack parameter $s$.
For the default uncertainty $s=1$, one of the seven events can be successfully attacked, resulting in a score of $f(\vec{x}^a;\vec{\theta}) \ll 1 $ while for the remaining six events no solution with a changed classification is found and the score remains close to $1$.
When scanning the events as a function of the attack parameter, it is found that relatively small uncertainties $s< 1 $ allow changing the classification for that one event, while the other six require substantially larger uncertainties of $s\approx 10$ (corresponding to the change in classification).
Such large uncertainties of roughly $\sigma_i \simeq \SI{100}{\percent} x_i $ are experimentally excluded.
When comparing simulated events, we find that misclassified $\nu_e$ events are typically attackable with attack parameters of $s \lesssim 1 $ similar to event 1, while correctly identified $\nu_\tau$ events require larger values $s\gg 1 $ similar to the other six events. This observation is consistent with the background estimation in~\cite{IceCube:2024nhk} of $0.5$ events. As a result, we see that the robustness of the network decision can be assessed by scanning the attack parameter, and we see a clear distinction between misclassified and correctly classified events. This information can be exploited on the one hand as an independent verification of analysis results and estimated uncertainties that depend on classification tasks or even to improve the quality and purity of the data selection itself.

\section{B-Jet Tagging with CMS Open Data}

In order to test how generalizable the above findings are, we apply \minifool to a very different test case in particle physics using data from the Compact Muon Solenoid (CMS) experiment~\cite{bib.cms_1, bib.cms_2}. We consider a standard classification task, which is the identification of heavy quarks that are produced in particle collisions.

The CMS detector is a multi-purpose detector for electrons, muons, photons, and both charged and neutral hadrons that are produced in proton-proton or heavy-ion collisions at the CERN Large Hadron Collider (LHC). It uses a global particle-flow (PF) algorithm~\cite{bib.particle_flow} to reconstruct all particles in an event.
For this, data from multiple subsystems are combined: the tracking detectors, the electromagnetic and hadron calorimeters, and the muon detectors.
 The reconstructed particles are then used to reconstruct the global kinematics of the event. Here, so-called jets are reconstructed~\cite{bib.anti_kt_1,bib.anti_kt_2,bib.anti_kt_3}.
 Jets are collimated sprays of particles that originate from the hadronization process of produced quarks and gluons~\cite{bib.jets_lhc}.

The focus of our study is so-called b-tagging, which is the identification of jets induced by a b-quark and their separation from the background which consists of jets originating from lighter quarks and gluons. As input, we use a publicly available simulated event sample from the CMS open data portal~\cite{bib.open_data}.
The simulated events originate from top-quark pair decay into W bosons and b-quarks.
The events are preselected for jet topologies and kinematic thresholds.
These jets are identified (tagged) by DeepJet, which is a deep neural network that has been developed by the CMS Collaboration~\cite{bib.deepjet_1,bib.deepjet_2}.
Inputs to this network are more than \num{600} quantities from various sources. These quantities include global characterizing values of the jet, such as momentum, orientation, and the number of particle constituents. Furthermore, numerous low-level features for the individual constituents are included, e.g.\ relative geometrical and kinematic values. The included values are both floating point as well as integer numbers, such as the multiplicity of particles or simple flags such as identified photons. The output of DeepJet for each tested jet is the probabilities $P$ for six different classes of jet-inducing particles. For this study, we compress this information into two classes: jets originating from b-quarks, denoted $P(B)$, and not from b-quarks, $P(\bar{B})$. The dataset is fully labeled on the simulated truth information, allowing for supervised training of DeepJet. In total, \num{10000000} jets were used for training, split into \num{8000000} for the training and \num{2000000} for the validation step, respectively. The inference is performed on \num{1000000} jets. Normalization of the input data is handled by internal batch normalization layers in the model~\cite{bib.batch_norm}.

\begin{figure}[htbp]
  \centering
  \includegraphics[width=.99\columnwidth]{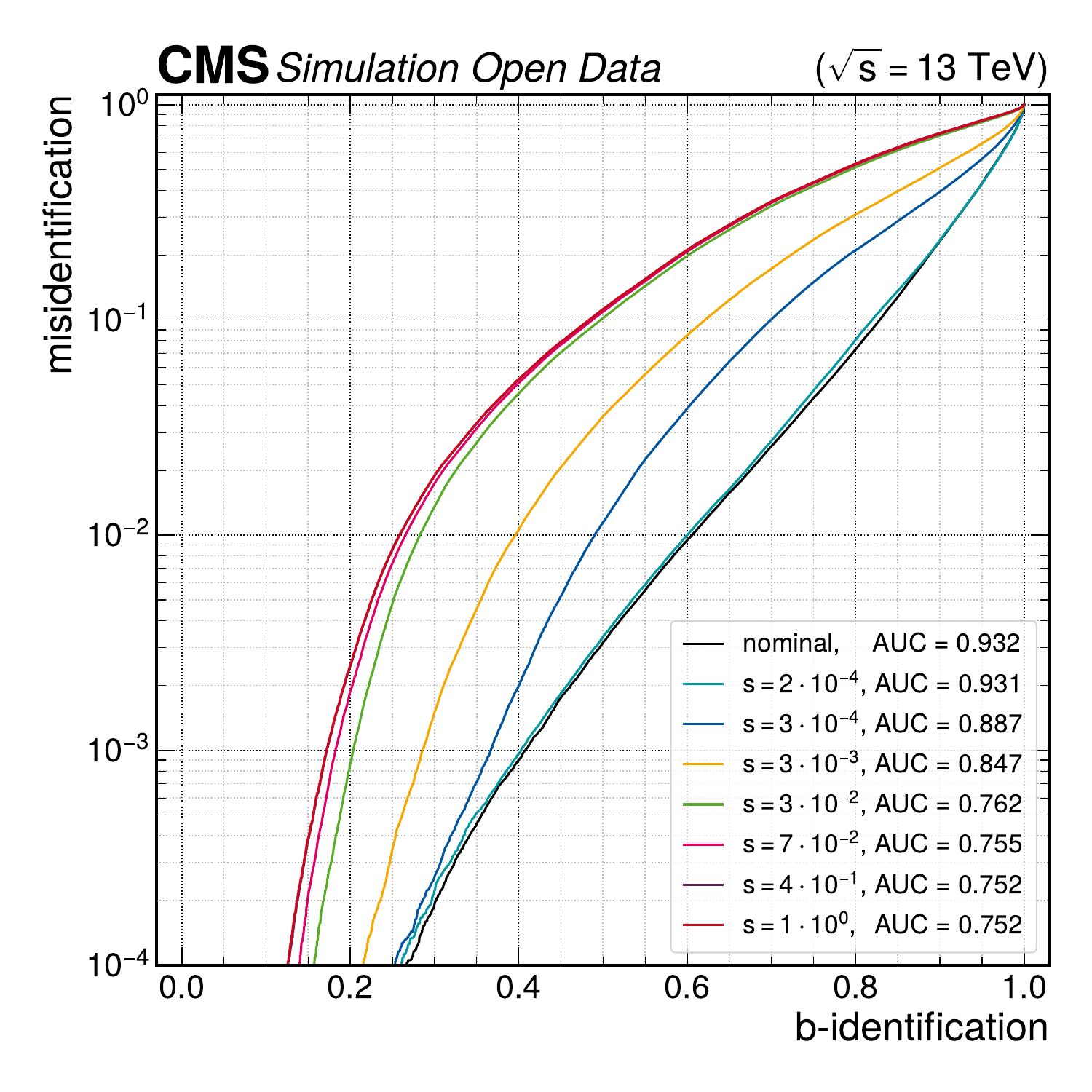}
  \caption{ROC curves for the jet-tagging performance for the full data set (see text). Shown is the nominally trained model and the results of perturbed data with different attack parameters.}
  \label{figure.cms_roc}
\end{figure}

Prior to the application of \minifool, we train DeepJet with the unperturbed (nominal) dataset.
To establish a baseline, we perform a nominal inference on the unperturbed test dataset. 
The performance is evaluated using a Receiver Operating Characteristic (ROC) curve~\cite{bib.roc_1, bib.roc_2} that shows the sliding cut of a discriminator. 
Generally, a discriminator between classes $X$ and $Y$ is defined as
\begin{align}
  \texttt{XvsY} = \frac{P(X)}{P(X) + P(Y)}.
\end{align}
In our simplified case of two classes, we can identify $\texttt{XvsY}~=~\texttt{BvsAll}~=~P(B)$.
To quantify the performance, the Area Under the Curve (AUC) is used as a measure\footnote{Note that we use the AUC convention of the CMS jet tagging group, where the used AUC is given by the area above the curve.}.
The nominal performance of the DeepJet model is shown by the black line in~\cref{figure.cms_roc}, yielding an AUC of \num{0.932}.

In the following, the impact of the \minifool attack 
on the decision-making of our nominal-trained DeepJet model is studied.
Each iteration is computationally expensive. Given the large size of the event samples, we perform \minifool with only ten iterations on the full dataset. We aim to maximize the change in the initial classification. For this purpose, we start from the total metric~\cref{equation.cost_function}, identify $g~=~0$, $\beta~=~1$, and use the adapted $L_2$ metric in~\cref{equation.eta} with
\begin{equation}
  \sigma_i = a_i \cdot s.
\end{equation}
The scaling variable $a_i$ normalizes the inputs, while $s$ is the attack parameter, as above.
In contrast to the previous applications, the nominal uncertainties $\vec{\sigma}^0$ of the more than 600 different input quantities are difficult to quantify. 
For simplicity of this test case, it is generally set to unity and varied using one global attack parameter.

The evaluation of DeepJet with perturbed inputs is shown in \cref{figure.cms_roc} for attack parameters of $\mathcal{O}(10^{-4}) - \mathcal{O}(10^{0})$.
While the performance of DeepJet with $s=\num{2e-4}$ recovers almost the nominal performance, the first degradation in performance can be observed for an attack parameter of \num{3e-4} with a corresponding AUC of \num{0.887}. Generating more severe input perturbations using higher attack parameters leads to more severe performance degradations. 
The additional degradations incurred beyond that caused by 
$s=\num{3e-2}$ are not as substantial as the degradations
below.
As expected for the proposed cost function, the dependency on $s$ is highly non-linear.
We show this by setting the attack parameter to unity, implying that the uncertainty is on the same order of magnitude as the input itself. Using this artificially high uncertainty, we can investigate the limit of degradation and find a convergence to an AUC of \num{0.752}.
This leads to the conclusion that 
DeepJet's classifications are  marginally affected for perturbations of its inputs that are smaller than 
\SI{0.02}{\%} of the normalized inputs.
Note, that the here used uncertainty model of simply scaling the input observables' values does not allow for a physical interpretation of  the $s$ value's range. The test shown here can be considered as a simple stress test that could be improved  with a more realistic uncertainty model.

\begin{figure}[htbp]
  \centering
  \includegraphics[width=.99\columnwidth]{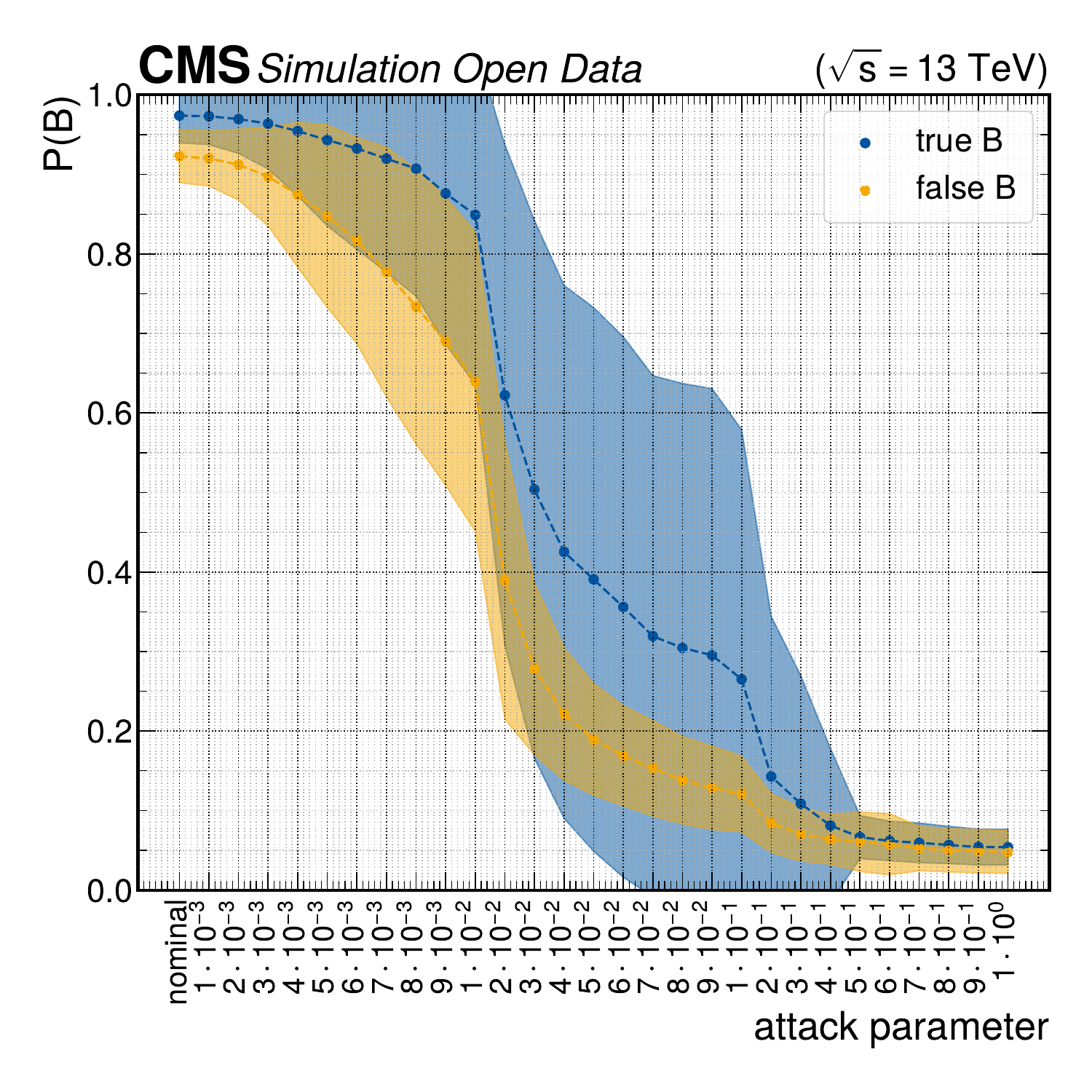}
  \caption{Network output as a function of the attack parameter $s$. The jets used here resemble a subset of the original test set (see text). By requiring an initial confidence of at least \SI{87}{\%}, we select \num{200} jets, split equally between correctly (blue) and incorrectly (orange) classified jets. The dotted curve shows the mean of the output $P(B)$, while the shaded region represents the corresponding standard deviation.}
  \label{figure.cms_score_tracking}
\end{figure}

As a next study, we investigate \num{200} randomly selected
jets of the class $P(B)$ which were classified with a confidence $P\ge \SI{87}{\percent}$ by the nominal model. Out of these, half are chosen from correctly classified $B$ jets (true), while the other half is misclassified as $B$ (false).
For these jets, we scan the robustness against the attack parameters $s$ similar to the previous test cases. For each tested $s$, we ensure full convergence of \minifool by allowing for up to \num{1000} iterations. \cref{figure.cms_score_tracking} shows the result of the scan.
For this subset of highly confident classification, we see the largest change in the classification shifted to larger attack parameters.
Furthermore, we reproduce the findings of
the previous test cases.
Increasing $s$ shows higher robustness for initially correctly classified jets compared to initially incorrectly classified ones, despite a similar high confidence of selected data.


\section{Discussion}

Adversarial attacks are being established as a powerful tool in the evaluation of physics analyses in the field of elementary particle and astroparticle physics.
However, as a challenge, typical algorithms for adversarial attacks do not preserve constraints imposed by physics unless supplemented by specific custom modifications.
In this paper, we present the new algorithm \minifool that combines experimental uncertainties with the goal of changing classifications by minimal adversarial perturbations of the input data. 
The algorithm
is applied to individual events and optimizes the minimum change according to the uncertainty of the input data and the changed target score. Depending on the assumed uncertainties, the attacks are successful or unsuccessful in case the required changes to the input are too costly.
We present three test cases: classification of MNIST numbers, identification of $\nu_\tau $ in the IceCube Neutrino Observatory, and Classification of tagged jets in the CMS experiment.
In all cases, we consistently find that the algorithm is capable of
testing classifications of experimental data. Furthermore, it allows
distinguishing between correctly classified and misclassified data by scanning the success of the attack as a function of the scale of the assumed uncertainty. 

Looking ahead, we see that the inclusion of physical boundary conditions into an adversarial attack enhances the possibilities of quantitatively evaluating the result from such an attack. For future applications, more complex uncertainty models need to be applied to achieve a more realistic application of experimental constraints.
While currently we have focused on testing decisions of pre-trained networks, the application within the training process also needs to be studied. However, here, the speed of the relatively slow numerical minimization process by the algorithm would currently limit the applicability to relatively small data samples.
As a further future aspect,
the question of applying a similar algorithm to regression tasks arises.
Here, the metric of a target score needs to be replaced by a metric that describes a significant deviation 
from the original regression result.


\backmatter

\bmhead{Supplementary information}

\bmhead{Acknowledgements}

We kindly acknowledge the permission to use simulated data from the IceCube and CMS Collaborations for this study. Particularly, we thank Doug Cowen for providing the classification network of IceCube's $\nu_\tau $ search.
This work has been conducted within the context of the \emph{AISafety} Project, that has been funded within the funding line
\emph{Software und Algorithmen zur Erforschung von Universum und Materie (ErUM) mit Schwerpunkt auf Künstlicher Intelligenz und Maschinellem Lernen} by the \emph{Bundesministerium f\"ur Bildung und Forschung (BMBF) of Germany}.

\bmhead{Author contributions}
The presented work has been developed as a collaborative effort by all authors within the \emph{AISafety} Project. 
O.J., P.A.J., P.S., M.T., and U.W. have contributed to the implementation of the code. The manuscript has been prepared by O.J., P.A.J., P.S., C.W., and U.W..
All authors have reviewed and contributed to the manuscript.

\bmhead{Declarations}
The authors declare no competing interests.

\bmhead{Code availability}

The code is available as both a TensorFlow and a PyTorch implementation
as open source on GitHub~\cite{git:minifool}.

\bmhead{Data availability}
The MNIST data set can be accessed through the used software packages TensorFlow and PyTorch.
The CMS data is publicly available through the CERN Open data portal.~\cite{bib.open_data}.
The IceCube simulation data is scheduled for a data release through the IceCube portal for public data \cite{icecube:open_data}.




\bibliography{biblio}

@misc{git:minifool,
    title  = {{MiniFool}},
    note   = {GitHub repository},
    url    = {https://github.com/ojanik/MiniFool},
    year   = {2025},
    author = {Oliver Janik and others},
}

@misc{masterjanik,
    title  = {{Using Adversarial Attacks to Fool IceCube’s Deep Neural Networks}},
    url    = {https://www.institut3b.physik.rwth-aachen.de/global/show_document.asp?id=aaaaaaaacgjxigu},
    author = {Oliver Janik},
    year   = {2023},
    note   = {Master Thesis, RWTH Aachen University},
}

@misc{icecube:open_data, 
    author = {{The IceCube Collaboration}},
    title  = {{Data Releases}},
    year   = {},
    note   = {IceCube portal for public data releases},
    url    = {https://icecube.wisc.edu/science/data-releases/},
}

@book{Erdmann:2021jbm,
    author = {Erdmann, Martin and Glombitza, Jonas and Kasieczka, Gregor and Klemradt, Uwe},
    title  = {{Deep Learning for Physics Research}},
    doi    = {10.1142/12294},
    month  = {2},
    year   = {2021},
}

@article{IceCube:2024nhk,
    author        = {{The IceCube Collaboration}},
    title         = {{Observation of Seven Astrophysical Tau Neutrino Candidates with IceCube}},
    eprint        = {2403.02516},
    archivePrefix = {arXiv},
    primaryClass  = {astro-ph.HE},
    doi           = {10.1103/PhysRevLett.132.151001},
    journal       = {Phys. Rev. Lett.},
    volume        = {132},
    number        = {15},
    pages         = {151001},
    year          = {2024},
}

@article{IceCube:2022der,
    author        = {{The IceCube Collaboration}},
    title         = {{Evidence for neutrino emission from the nearby active galaxy NGC 1068}},
    eprint        = {2211.09972},
    archivePrefix = {arXiv},
    primaryClass  = {astro-ph.HE},
    doi           = {10.1126/science.abg3395},
    journal       = {Science},
    volume        = {378},
    number        = {6619},
    pages         = {538--543},
    year          = {2022},
}

@article{IceCube:2023ame,
    author        = {{The IceCube Collaboration}},
    title         = {{Observation of high-energy neutrinos from the Galactic plane}},
    eprint        = {2307.04427},
    archivePrefix = {arXiv},
    primaryClass  = {astro-ph.HE},
    doi           = {10.1126/science.adc9818},
    journal       = {Science},
    volume        = {380},
    number        = {6652},
    pages         = {adc9818},
    year          = {2023},
}

@article{PhysRevLett.134.021001,
    title     = {{Inference of the Mass Composition of Cosmic Rays with Energies from ${10}^{18.5}$ to ${10}^{20}\text{ }\text{ }\mathrm{eV}$ Using the Pierre Auger Observatory and Deep Learning}},
    author    = {{Pierre Auger Collaboration}},
    journal   = {Phys. Rev. Lett.},
    volume    = {134},
    issue     = {2},
    pages     = {021001},
    numpages  = {10},
    year      = {2025},
    month     = {Jan},
    publisher = {American Physical Society},
    doi       = {10.1103/PhysRevLett.134.021001}
}

@article{hashemi2024ultra,
    title     = {{Ultra-high-granularity detector simulation with intra-event aware generative adversarial network and self-supervised relational reasoning}},
    author    = {Hashemi, Baran and Hartmann, Nikolai and Sharifzadeh, Sahand and Kahn, James and Kuhr, Thomas},
    journal   = {Nature Communications},
    volume    = {15},
    number    = {1},
    pages     = {4916},
    year      = {2024},
    publisher = {Nature Publishing Group UK London},
    doi       = {10.1038/s41467-024-49104-4}
}

@article{lai2022explainable,
    author  = {Lai, Yue Shi and Neill, Duff and P{\l}osko{\'n}, Mateusz and Ringer, Felix},
    title   = {{Explainable machine learning of the underlying physics of high-energy particle collisions}},
    journal = {Physics Letters B},
    volume  = {829},
    pages   = {137055},
    year    = {2022},
    issn    = {0370-2693},
    doi     = {10.1016/j.physletb.2022.137055}
}

@article{lagan,
   title     = {{Learning Particle Physics by Example: Location-Aware Generative Adversarial Networks for Physics Synthesis}},
   volume    = {1},
   issn      = {2510-2044},
   doi       = {10.1007/s41781-017-0004-6},
   number    = {1},
   journal   = {Computing and Software for Big Science},
   publisher = {Springer Science and Business Media LLC},
   author    = {de Oliveira, Luke and Paganini, Michela and Nachman, Benjamin},
   year      = {2017}
}

@article{IceCube:2016zyt,
    author        = {{The IceCube Collaboration}},
    title         = {{The IceCube Neutrino Observatory: Instrumentation and Online Systems}},
    eprint        = {1612.05093},
    archivePrefix = {arXiv},
    primaryClass  = {astro-ph.IM},
    doi           = {10.1088/1748-0221/12/03/P03012},
    journal       = {JINST},
    volume        = {12},
    number        = {03},
    pages         = {P03012},
    year          = {2017},
    note          = {[Erratum: JINST 19, E05001 (2024)]},
}

@article{deFavereau:2013fsa,
    author        = {de Favereau, J. and Delaere, C. and Demin, P. and Giammanco, A. and Lemaître, V. and Mertens, A. and Selvaggi, M.},
    title         = {{DELPHES 3, A modular framework for fast simulation of a generic collider experiment}},
    eprint        = {1307.6346},
    archivePrefix = {arXiv},
    primaryClass  = {hep-ex},
    doi           = {10.1007/JHEP02(2014)057},
    journal       = {JHEP},
    volume        = {02},
    pages         = {057},
    year          = {2014},
}

@article{deng2012mnist,
    author  = {Deng, Li},
    journal = {IEEE Signal Processing Magazine}, 
    title   = {{The MNIST Database of Handwritten Digit Images for Machine Learning Research [Best of the Web]}},
    year    = {2012},
    volume  = {29},
    number  = {6},
    pages   = {141-142},
    doi     = {10.1109/MSP.2012.2211477},
}

@article{xu2020adversarial,
    title     = {{Adversarial attacks and defenses in images, graphs and text: A review}},
    author    = {Xu, Han and Ma, Yao and Liu, Hao-Chen and Deb, Debayan and Liu, Hui and Tang, Ji-Liang and Jain, Anil K},
    journal   = {International journal of automation and computing},
    volume    = {17},
    pages     = {151--178},
    year      = {2020},
    publisher = {Springer},
    doi       = {10.1007/s11633-019-1211-x},
}

@misc{lecun2010mnist,
    title   = {{MNIST handwritten digit database}},
    author  = {LeCun, Yann and Cortes, Corinna and Burges, CJ},
    journal = {ATT Labs [Online]},
    volume  = {2},
    year    = {2010},
    url     = {http://yann.lecun.com/exdb/mnist},
}

@misc{abadi2016tensorflow,
    title         = {{TensorFlow: A system for large-scale machine learning}},
    author        = {Martín Abadi and Paul Barham and Jianmin Chen and Zhifeng Chen and Andy Davis and Jeffrey Dean and Matthieu Devin and Sanjay Ghemawat and Geoffrey Irving and Michael Isard and Manjunath Kudlur and Josh Levenberg and Rajat Monga and Sherry Moore and Derek G. Murray and Benoit Steiner and Paul Tucker and Vijay Vasudevan and Pete Warden and Martin Wicke and Yuan Yu and Xiaoqiang Zheng},
    year          = {2016},
    eprint        = {1605.08695},
    archivePrefix = {arXiv},
    primaryClass  = {cs.DC},
    doi           = {10.48550/arXiv.1605.08695}, 
}

@misc{tensorflow2015-whitepaper,
    title  = {{TensorFlow: Large-Scale Machine Learning on Heterogeneous Systems}},
    url    = {https://www.tensorflow.org/},
    note   = {Software available from tensorflow.org},
    author = {Martín Abadi and Ashish Agarwal and Paul Barham and Eugene Brevdo and Zhifeng Chen and Craig Citro and Greg S. Corrado and Andy Davis and Jeffrey Dean and Matthieu Devin and Sanjay Ghemawat and Ian Goodfellow and Andrew Harp and Geoffrey Irving and Michael Isard and Yangqing Jia and Rafal Jozefowicz and Lukasz Kaiser and Manjunath Kudlur and Josh Levenberg and Dan Mané and Rajat Monga and Sherry Moore and Derek Murray and Chris Olah and Mike Schuster and Jonathon Shlens and Benoit Steiner and Ilya Sutskever and Kunal Talwar and Paul Tucker and Vincent Vanhoucke and Vijay Vasudevan and Fernanda Viégas and Oriol Vinyals and Pete Warden and Martin Wattenberg and Martin Wicke and Yuan Yu and Xiaoqiang Zheng},
    year   = {2015},
}

@misc{paszke2017automatic,
    title     = {{Automatic differentiation in PyTorch}},
    author    = {Paszke, Adam and Gross, Sam and Chintala, Soumith and Chanan, Gregory and Yang, Edward and DeVito, Zachary and Lin, Zeming and Desmaison, Alban and Antiga, Luca and Lerer, Adam},
    booktitle = {NIPS-W},
    year      = {2017},
    url       = {https://openreview.net/pdf?id=BJJsrmfCZ},
}

@article{paszke2019pytorch,
      title         = {{PyTorch: An Imperative Style, High-Performance Deep Learning Library}},
      author        = {Adam Paszke and Sam Gross and Francisco Massa and Adam Lerer and James Bradbury and Gregory Chanan and Trevor Killeen and Zeming Lin and Natalia Gimelshein and Luca Antiga and Alban Desmaison and Andreas Köpf and Edward Yang and Zach DeVito and Martin Raison and Alykhan Tejani and Sasank Chilamkurthy and Benoit Steiner and Lu Fang and Junjie Bai and Soumith Chintala},
      year          = {2019},
      eprint        = {1912.01703},
      archivePrefix = {arXiv},
      primaryClass  = {cs.LG},
      doi           = {10.48550/arXiv.1912.01703}, 
}

@article{Goodfellow:2014rpb,
    title         = {{Explaining and Harnessing Adversarial Examples}},
    author        = {Ian J. Goodfellow and Jonathon Shlens and Christian Szegedy},
    year          = {2015},
    eprint        = {1412.6572},
    archivePrefix = {arXiv},
    primaryClass  = {stat.ML},
    doi           = {10.48550/arXiv.1412.6572}, 
}

@inproceedings{Deepfoolpaper,
    author    = {Moosavi-Dezfooli, Seyed-Mohsen and Fawzi, Alhussein and Frossard, Pascal},
    booktitle = {2016 IEEE Conference on Computer Vision and Pattern Recognition (CVPR)},
    title     = {{DeepFool: A Simple and Accurate Method to Fool Deep Neural Networks}},
    year      = {2016},
    pages     = {2574-2582},
    doi       = {10.1109/CVPR.2016.282},
}

@article{Flek:2025ecg,
    title         = {{Enforcing Fundamental Relations via Adversarial Attacks on Input Parameter Correlations}},
    author        = {Timo Saala and Lucie Flek and Alexander Jung and Akbar Karimi and Alexander Schmidt and Matthias Schott and Philipp Soldin and Christopher Wiebusch},
    year          = {2025},
    eprint        = {2501.05588},
    archivePrefix = {arXiv},
    primaryClass  = {cs.LG},
    doi           = {10.48550/arXiv.2501.05588}, 
}

@article{Madry:2017tvh,
    title         = {{Towards Deep Learning Models Resistant to Adversarial Attacks}},
    author        = {Aleksander Madry and Aleksandar Makelov and Ludwig Schmidt and Dimitris Tsipras and Adrian Vladu},
    year          = {2019},
    eprint        = {1706.06083},
    archivePrefix = {arXiv},
    primaryClass  = {stat.ML},
    doi           = {10.48550/arXiv.1706.06083}, 
}

@inproceedings{carlini2017,
    author    = {Carlini, Nicholas and Wagner, David},
    booktitle = {2017 IEEE Symposium on Security and Privacy (SP)},
    title     = {{Towards Evaluating the Robustness of Neural Networks}},
    year      = {2017},
    pages     = {39-57},
    doi       = {10.1109/SP.2017.49},
}

@article{szegedy2013intriguing,
    title         = {{Intriguing properties of neural networks}},
    author        = {Christian Szegedy and Wojciech Zaremba and Ilya Sutskever and Joan Bruna and Dumitru Erhan and Ian Goodfellow and Rob Fergus},
    year          = {2014},
    eprint        = {1312.6199},
    archivePrefix = {arXiv},
    primaryClass  = {cs.CV},
    doi           = {10.48550/arXiv.1312.6199}, 
}

@phdthesis{Stein:991721,
    author            = {Stein, Annika},
    othercontributors = {Schmidt, Alexander and Krämer, Michael},
    title             = {{Novel jet flavour tagging algorithms exploiting adversarial deep learning techniques with efficient computing methods and preparation of open data for robustness studies}},
    school            = {RWTH Aachen University},
    type              = {Dissertation},
    address           = {Aachen},
    publisher         = {RWTH Aachen University},
    reportid          = {RWTH-2024-07840},
    pages             = {1 Online-Ressource : Illustrationen},
    year              = {2024},
    note              = {Veröffentlicht auf dem Publikationsserver der RWTH Aachen University; Dissertation, RWTH Aachen University, 2024},
    cin               = {133920 ; 133910 / 130000},
    ddc               = {530},
    cid               = {$I:(DE-82)133920_20180228$ / $I:(DE-82)130000_20140620$},
    typ               = {PUB:(DE-HGF)11},
    doi               = {10.18154/RWTH-2024-07840},
}

@article{IceCube:2020fpi,
    author        = {{The IceCube Collaboration}},
    title         = {{Detection of astrophysical tau neutrino candidates in IceCube}},
    eprint        = {2011.03561},
    archivePrefix = {arXiv},
    primaryClass  = {hep-ex},
    doi           = {10.1140/epjc/s10052-022-10795-y},
    journal       = {Eur. Phys. J. C},
    volume        = {82},
    number        = {11},
    pages         = {1031},
    year          = {2022},
}

@article{IceCube:2013low,
    author        = {{The IceCube Collaboration}},
    title         = {{Evidence for High-Energy Extraterrestrial Neutrinos at the IceCube Detector}},
    eprint        = {1311.5238},
    archivePrefix = {arXiv},
    primaryClass  = {astro-ph.HE},
    doi           = {10.1126/science.1242856},
    journal       = {Science},
    volume        = {342},
    pages         = {1242856},
    year          = {2013},
}

@article{Learned:1994wg,
    author        = {Learned, John G. and Pakvasa, Sandip},
    title         = {{Detecting tau-neutrino oscillations at PeV energies}},
    eprint        = {hep-ph/9405296},
    archivePrefix = {arXiv},
    reportNumber  = {UH-511-799-94, DUMAND-3-94},
    doi           = {10.1016/0927-6505(94)00043-3},
    journal       = {Astropart. Phys.},
    volume        = {3},
    pages         = {267--274},
    year          = {1995},
}

@article{IceCube:2015vkp,
    author        = {{The IceCube Collaboration}},
    title         = {{Search for Astrophysical Tau Neutrinos in Three Years of IceCube Data}},
    eprint        = {1509.06212},
    archivePrefix = {arXiv},
    primaryClass  = {astro-ph.HE},
    doi           = {10.1103/PhysRevD.93.022001},
    journal       = {Phys. Rev. D},
    volume        = {93},
    number        = {2},
    pages         = {022001},
    year          = {2016},
}

@article{Simonyan:2014cmh,
    title         = {{Very Deep Convolutional Networks for Large-Scale Image Recognition}},
    author        = {Karen Simonyan and Andrew Zisserman},
    year          = {2015},
    eprint        = {1409.1556},
    archivePrefix = {arXiv},
    primaryClass  = {cs.CV},
    doi           = {10.48550/arXiv.1409.1556},
}

@article{Mosteller01101948,
    author    = {Frederick Mosteller},
    title     = {{Questions and Answers}},
    journal   = {The American Statistician},
    volume    = {2},
    number    = {5},
    pages     = {30--31},
    year      = {1948},
    publisher = {ASA Website},
    doi       = {10.1080/00031305.1948.10483405},
}

@article{bib.open_data, 
    author = {{The CMS Collaboration}},
    title  = {{Simulated dataset  TTToHadronic\_TuneCP5\_13TeV-powheg-pythia8 in MINIAODSIM format for 2016 collision data}},
    year   = {2024},
    note   = {[CERN Open Data Portal]},
    doi    = {10.7483/OPENDATA.CMS.G6CE.1ITV},
}

@article{bib.cms_1,
    author  = {{The CMS Collaboration}},
    title   = {{The CMS Experiment at the CERN LHC}},
    doi     = {10.1088/1748-0221/3/08/S08004},
    journal = {JINST},
    volume  = {3},
    pages   = {S08004},
    year    = {2008},
}

@article{bib.cms_2,
    author  = {{The CMS Collaboration}},
    title   = {{Development of the CMS detector for the CERN LHC Run 3}},
    doi     = {10.1088/1748-0221/19/05/P05064},
    journal = {JINST},
    volume  = {19},
    pages   = {P05064},
    year    = {2024},
}

@book{bib.jets_lhc,
    title     = {{Jet Physics at the LHC: The Strong Force beyond the TeV Scale}},
    author    = {Klaus Rabbertz},
    year      = {2017},
    publisher = {{Springer Cham}},
    series    = {{Springer Tracts in Modern Physics}},
    doi       = {10.1007/978-3-319-42115-5},
    isbn      = {978-3-319-42115-5},
    volume    = {268},
}

@article{bib.anti_kt_1,
    title     = {{The anti-$k_t$ jet clustering algorithm}},
    volume    = {2008},
    issn      = {1029-8479},
    doi       = {10.1088/1126-6708/2008/04/063},
    number    = {04},
    journal   = {Journal of High Energy Physics},
    publisher = {Springer Science and Business Media LLC},
    author    = {Cacciari, Matteo and Salam, Gavin P and Soyez, Gregory},
    year      = {2008},
    pages     = {063–063},
}

@article{bib.anti_kt_2,
    title     = {{Successive combination jet algorithm for hadron collisions}},
    volume    = {48},
    issn      = {0556-2821},
    doi       = {10.1103/physrevd.48.3160},
    number    = {7},
    journal   = {Physical Review D},
    publisher = {American Physical Society (APS)},
    author    = {Ellis, Stephen D. and Soper, Davison E.},
    year      = {1993},
    pages     = {3160–3166},
}

@article{bib.anti_kt_3,
    title   = {{Longitudinally-invariant $k_\perp$-clustering algorithms for hadron-hadron collisions}},
    journal = {Nuclear Physics B},
    volume  = {406},
    number  = {1},
    pages   = {187-224},
    year    = {1993},
    issn    = {0550-3213},
    doi     = {10.1016/0550-3213(93)90166-M},
    author  = {S. Catani and Yu.L. Dokshitzer and M.H. Seymour and B.R. Webber},
}

@article{bib.particle_flow,
    doi     = {10.1088/1748-0221/12/10/P10003},
    year    = {2017},
    volume  = {12},
    number  = {10},
    pages   = {P10003},
    author  = {{The CMS Collaboration}},
    title   = {{Particle-flow reconstruction and global event description with the CMS detector}},
    journal = {Journal of Instrumentation},
}

@article{bib.deepjet_1,
    title     = {{Jet flavour classification using DeepJet}},
    volume    = {15},
    issn      = {1748-0221},
    doi       = {10.1088/1748-0221/15/12/p12012},
    number    = {12},
    journal   = {Journal of Instrumentation},
    publisher = {IOP Publishing},
    author    = {Bols, E. and Kieseler, J. and Verzetti, M. and Stoye, M. and Stakia, A.},
    year      = {2020},
    month     = {dec},
    pages     = {P12012},
}

@misc{bib.deepjet_2,
    author = {DL4Jets},
    title  = {{DeepJet: Repository for training and evaluation of deep neural networks for jet identification}},
    year   = {2020},
    url    = {https://github.com/DL4Jets/DeepJet},
    note   = {[Accessed: 02.09.2025]},
}

@misc{bib.batch_norm,
    title         = {{Batch Normalization: Accelerating Deep Network Training by Reducing Internal Covariate Shift}}, 
    author        = {Sergey Ioffe and Christian Szegedy},
    year          = {2015},
    eprint        = {1502.03167},
    archivePrefix = {arXiv},
    primaryClass  = {cs.LG},
    doi           = {10.48550/arXiv.1502.03167}, 
}

@article{bib.roc_1,
    title   = {{The use of the area under the ROC curve in the evaluation of machine learning algorithms}},
    journal = {Pattern Recognition},
    volume  = {30},
    number  = {7},
    pages   = {1145-1159},
    year    = {1997},
    issn    = {0031-3203},
    doi     = {10.1016/S0031-3203(96)00142-2},
    author  = {Andrew P. Bradley},
}

@article{bib.roc_2,
    title   = {{An introduction to ROC analysis}},
    journal = {Pattern Recognition Letters},
    volume  = {27},
    number  = {8},
    pages   = {861-874},
    year    = {2006},
    note    = {ROC Analysis in Pattern Recognition},
    issn    = {0167-8655},
    doi     = {10.1016/j.patrec.2005.10.010},
    author  = {Tom Fawcett},
}

@article{Stein:2022nvf,
    author        = {Stein, Annika and Coubez, Xavier and Mondal, Spandan and Novak, Andrzej and Schmidt, Alexander},
    title         = {{Improving Robustness of Jet Tagging Algorithms with Adversarial Training}},
    eprint        = {2203.13890},
    archivePrefix = {arXiv},
    primaryClass  = {physics.data-an},
    doi           = {10.1007/s41781-022-00087-1},
    journal       = {Comput. Softw. Big Sci.},
    volume        = {6},
    number        = {1},
    pages         = {15},
    year          = {2022}
}

@misc{kingma2017adammethodstochasticoptimization,
      title={Adam: A Method for Stochastic Optimization}, 
      author={Diederik P. Kingma and Jimmy Ba},
      year={2017},
      eprint={1412.6980},
      archivePrefix={arXiv},
      primaryClass={cs.LG},
      url={https://arxiv.org/abs/1412.6980}
}

\end{document}